\pdfoutput=1

\documentclass[11pt]{article}

\usepackage{acl}

\usepackage{times}
\usepackage{latexsym}

\usepackage[T1]{fontenc}

\usepackage[utf8]{inputenc}

\usepackage{microtype}

\usepackage{graphicx, subcaption}

\usepackage{tablefootnote}
\newif\ifcomment\commenttrue
\ifcomment
    \newcommand{\pinaforecomment}[3]{\colorbox{#1}{\parbox{.8\linewidth}{#2: #3}}}
\else
    \newcommand{\pinaforecomment}[3]{}
\fi

\usepackage{enumitem}
\usepackage{hyperref}

\def\example A\Q#1\QA#2\SK#3\R#4 B{\noindent\fbox{\parbox{\linewidth}{\small{\texttt{\textbf{Question: }#1\\ \textbf{QA Answer: }#2\\ \textbf{Secret Answer: }#3\\ \textbf{Result: }#4}}}}\\}

\title{Secret-Keeping in Question Answering}

\author{Nathaniel W. Rollings\textsuperscript{*} \and Kent O'Sullivan\textsuperscript{*} \and Sakshum Kulshrestha\\
  Department of Computer Science\\
  University of Maryland \\
  \texttt{\{nrolling, osullik, sakshumk\}@umd.edu}\\
  \footnotesize * equal contribution by the authors}
  
\begin{document}
\maketitle
\begin{abstract}
Existing question answering research focuses on unanswerable questions
in the context of always providing an answer when a system can\dots
but what about cases where a system {\bf should not} answer a
question.
This can either be to protect sensitive users or sensitive
information.
Many models expose sensitive information under interrogation by an
adversarial user.
We seek to determine if it is possible to teach a question-answering
system to keep a specific fact secret.
We design and implement a proof-of-concept architecture and through
our evaluation determine that while possible, there are numerous
directions for future research to reduce system paranoia (false
positives), information leakage (false negatives) and extend the
implementation of the work to more complex problems with preserving
secrecy in the presence of information aggregation.
\end{abstract}

\section{Introduction} \label{intro}

\subsection{Protecting secrets and serving answers}

\textbf{Answering Questions.} It is evident from the survey completed by \citeauthor{Rogers2023} that the majority of work in Question Answering (QA) seeks to improve accuracy. For a long time, the focus has been maximizing the \textit{availability} of an answer given text. Work into the \textit{answerability} of questions is improving the \textit{integrity} of returned information. No work we have found has explicitly considered the \textit{confidentiality} of answers in question answering. The focus has been exhaustively on whether a QA system \textit{can} answer a question accurately. In this paper, we ask \textit{should} a question be answered accurately, and how do we prevent \textit{secret} information from being disclosed by a QA system? \citeauthor{Rogers2023} contend that work in QA systems is increasingly driven by commercial demand, emphasizing the impact of the assertion from \citeauthor{Roy2021} that the objectives of a commercial system may differ from the broader goal of building a QA system with complex and higher reasoning ability. The requirement to provide \textit{confidentiality} features in QA systems with access to sensitive corporate data is almost certain, and there has been a lack of work examining the problem. Worryingly, a 2022 study by \citeauthor{Jagielski2022} indicates that memorization of training data is more likely on recently observed examples in Large Language Models (LLMs). As the focus of QA shifts towards answer generation \cite{Baradaran2022}, systems like ChatGPT \cite{OpenAI2022} are increasingly likely to be deployed in a corporate setting. Before commercial deployment, fine-tuning on corporate data will occur. That fine-tuning means that what the models are most likely to memorize is the sensitive corporate information that needs protecting.

\textbf{Protecting Information. }Current approaches to protect against the leaking of secrets are inadequate. \textit{Censoring} secrets in the potential answer context are suboptimal. Work by \citeauthor{Song2019} in 2019 showed that censoring training data degrades performance and, in some cases, is reversible, exposing the sensitive material. Complete redaction is an option to protect secrets; however, a counter-factual analysis by \citeauthor{Kandpal2022} suggests that (unsurprisingly) the overall performance of a generative QA model drops when the context is redacted. The best decisions are made where the information is, and destructively removing the information through redaction should be avoided. A detailed exploration of related work is in section \ref{related}, but to build intuition about why we should care about this problem of whether a computer \textit{should} be able to answer every question, consider the following fictional scenario: 

\textbf{Motivation.} After years of complaints from Wile E. Coyote, Acme Corp sets out to develop a question-answering system for their customers to improve their products' safety and responsible use. Acme expects most users to ask \textit{how-to} questions before or \textit{first-aid} questions after using a product. To ensure that the system can answer as many questions as possible, they make available to it every official document ever written by Acme - technical reports, patents, product designs, investigations, reviews, meeting minutes etc. Initial testing is promising. The system can accurately answer most of the common questions like \textit{"How do I decrease the thrust on the Acme rocket skates"} and \textit{"How to treat burns from Acme little-giant firecracker"}. Unfortunately, it also happily provides answers containing proprietary (and dangerous) information. Testers found that asking \textit{"how do I build an Acme Self-Guided Aerial Bomb"} yields detailed instructions on how to build the device. Not wanting to give away confidential designs or be liable for a spate of DIY bomb-makers, Acme tried to remove the underlying information about the Acme Self-Guided Aerial Bomb. They found that redacting the references to the Bomb's construction also degraded the system's ability to answer questions about whether or not the Self Guided Aerial bomb is made of ethically sourced materials and what type of fuel needs to be put in it. So, how does Acme Corp stop their system from leaking damaging or dangerous information into the wrong hands while releasing other information? How do they maximize their ability to share information while protecting their secrets?




\textbf{QA Systems.} The generation of succinct answers to questions over increasingly heterogeneous modalities is achieved by \textit{question answering} (QA). 
QA systems seek to provide a direct answer to an information need posed by a user in natural language \cite{Roy2021}. QA systems can be characterized by their \textit{question input}, \textit{context input} and \textit{output}. Input questions can be \textit{information seeking}, where the user is trying to gain knowledge they do not yet have, or \textit{probing}, where the user is confirming what knowledge a system has \cite{Rogers2023}. The source of the material a QA system will answer questions about is called the \textit{context}. The context of a QA system is typically sourced from a structured knowledge base or an unstructured collection. Unstructured collections can cover any modality but are most heavily concentrated on unstructured text \cite{Roy2021}. Systems optimized for unstructured text are also known as \textit{reading comprehension} or \textit{machine reading} systems. Outputs of a QA system can be \textit{extractive} returning a span of text or knowledge base entry located within the context to satisfy the information need, \textit{generative}, creating a novel answer to the information requirement, \textit{multi-choice} selecting from a list of possibilities or \textit{categorical} for example, \textit{yes/no}  \cite{Rogers2023}.


\textbf{Evaluating QA Systems.} Current QA evaluation focuses on measuring the 'accuracy' of returned answers: did the provided answer satisfy the information requirement of the question with respect to the context? These measures are typically evaluated against a 'gold standard' dataset. These gold standards contain, at a minimum: contexts, questions and acceptable answers. \textit{Extractive, multi-choice and categorical} systems are typically measured using \textit{exact match accuracy}, \textit{F1 score} or one of a few specialized metrics outlined by \citeauthor{Baradaran2022}. \textit{Generative} systems also use specialized measures primarily based on n-gram comparisons; detail is available in work by \citeauthor{Baradaran2022}. The 2023 survey paper by \citeauthor{Rogers2023} gives a detailed overview of available datasets for evaluation. Most follow the gold-standard pattern with variations to the source of questions and context. Most relevant to protecting information is work exploring \textit{answerability}, the measure of whether or not a QA system is \textit{capable} of answering a given question. \citeauthor{Roy2021} offer three  common reasons for questions being unanswerable: \textit{low confidence} in the answer, the requested information being \textit{not in the context} and when \textit{the correct answer is null}. \citeauthor{Rajpurkar2018} published the \textit{SQUAD2.0} dataset in 2018, allowing systems to evaluate against unanswerable questions. Work by \citeauthor{Wallace2019} in 2019 shows how deliberately constructing adversarial questions can make questions unanswerable, and the \textit{QuAIL} dataset published in 2020 by \citeauthor{Rogers2020} expands further on answerability, noting that the systems they tested were rarely able to identify if a question was unanswerable with an accuracy greater than chance. All of these metrics focus on whether a system \textit{can} answer a question, but none of them addresses if an answer it provided \textit{should} have been provided.

\textbf{Research Question.} How can we implement a secret-keeping system capable of protecting secret information from disclosure without inflicting unacceptable degradation on the system's ability to provide non-secret answers to questions? 

\textbf{Organization.} The remainder of the paper gives key definitions, explains the \textit{design} of our system in section \ref{design}, our \textit{experiment}s in section \ref{experiment} and the \textit{results} in section \ref{evaluation}. Sections \ref{ethics}, \ref{related} and \ref{conclusion} outline the \textit{ethics}, \textit{related work} and \textit{future work}, respectively.\\ 


\textbf{Contributions}. Our key contributions are: 
\begin{itemize}[nosep,labelindent=0pt,itemindent=0pt,leftmargin=*]
    \item \textbf{Problem Definition.} We identify the gaps in assuring confidentiality in QA systems and introduce \textit{secret-keeping} as a solution.
    \item \textbf{Architecture.} We develop a flexible architecture that can be easily adapted to different question-answering systems to protect secret information from unauthorized disclosure. 
    \item \textbf{Evaluation} We develop evaluation metrics to assess the effectiveness of a secret-keeping model.
\end{itemize}

\subsection{What is a secret?} \label{definition}


 Before building systems that prevent the leaking of secret information, we need to define what a secret is, how we imagine users interacting with the system that can access secret and non-secret material, and what can go wrong:

\begin{itemize}[nosep,labelindent=0pt,itemindent=0pt,leftmargin=*]
    \item A \textbf{\textit{secret}} is an atomic fact or a relationship between entities that should not be disclosed. We define the metric of \textit{secrecy} as the proportion of answers containing secrets that were correctly identified by the secret-keeper. 
    \item \textbf{\textit{Interrogation}} is an adversarial interaction with the question-answering entity where questions are deliberately constructed to induce information leakage. An effective interrogator can introduce information leakage by  crafting adversarial in-band interactions with the system\footnote{See what happened to ChatGPT\cite{samczsun2022, Johnson2022}} or by employing out-of-band privacy attacks exploiting the memorization of training examples like those described by \citeauthor{Kandpal2022a} in their 2022 paper. 
    \item\textbf{\textit{Information leakage}} we use similarly to \citealt{FAIR2022}s use in 2022, referring to situations where the [question answering] agent reveals compromising information about its plan [secret]. We define the \textit{information leakage} metric as the proportion of answers containing secrets not identified by the secret-keeper (.i.e. False Negatives).
    \item \textbf{\textit{Paranoia}} describes when the QA agent refuses to answer a question that does not contain a secret. We define the \textit{paranoia} metric as the proportion of answers not containing secrets that were incorrectly identified as containing secrets by the secret-keeper (i.e. False Positives).
\end{itemize}



\section{Designing a system to keep secrets} \label{design}



\textbf{Design Principles.} Our review of the literature in section \ref{related} identifies several areas of related work that have informed the design of our system. Key influences are noted in our list of design principles used to develop the secret keeper approach below:\\ 
\begin{enumerate}[nosep,labelindent=0pt,itemindent=0pt,leftmargin=*]
    \item The system must \textbf{minimize} \textit{information leakage} as work in agent based systems reviewed in section \ref{related_agents} identifies as a priority.
    \item The system should \textbf{minimize} \textit{paranoia} so that it performs effectively as a QA system for non-secret information as described in section \ref{related_QA}.
    \item The system should not \textbf{destroy} context to prevent information loss as explored in section \ref{related_memorization}.
    \item The system should \textbf{generalize} to extractive, generative, multi-choice and categorical QA. 
    \item The sanitizing should be \textbf{invisible} to users, concealing omission balances ethics and efficacy as discussed in section \ref{ethics}.
\end{enumerate}

As an initial proof-of-concept, we deliberately limit our scope to extractive, machine-reading QA systems. Our design principles prioritize an approach that focuses on filtering the \textit{output} of a system rather than redacting or censoring the input. The filtering of output is referred to as \textit{sanitization} by \citeauthor{Carlini2019} in their 2019 paper, asserting that it is the best practice for processing sensitive and private data. We implement the following \textit{secret keeping architecture}:

\textbf{Secret Keeper Architecture.} We designed the output-sanitization architecture shown in figure \ref{fig:sanitize} to offer significant flexibility to a practitioner. By focusing exclusively on the output we make our system agnostic to the underlying model, QA method and context. The secret keeper only stops secrets from being disclosed. It determines if a QA answer is a secret by asking the same question the QA system is answering of its own \textit{secret} context. The \textit{secret} context contains \textit{only} secrets. The secret keeper takes uses the cosine similarity of the QA answer. If the similarity is >0.5, the QA answer is determined to be secret and the output is sanitized.


\section{Verifying that secrets are being kept} \label{experiment}
We approach the experimental design in three phases. 

%
\subsection{Phase 1. Baseline Assessment}\label{exp_baselines}
The purpose of our baseline assessment is to measure the performance of each of our QA models on an unmodified context without interference from the secret-keeper. The input questions and context are both sourced from the SQUAD Dev Set\footnote{\href{https://rajpurkar.github.io/SQuAD-explorer/explore/1.1/dev/}{Squad Dev set}} \cite{Rajpurkar2016}. The secrets are a non-disjoint subset of the SQUAD Dev Set. The context has not been redacted and the secret keeper is not active. We measure the Accuracy of the QA using the SQUAD gold standard answers, and the \textit{information leakage} and \textit{paranoia} of the system using the gold standard answers and membership of the secret subgroup. The models we use are: 
\begin{itemize}[nosep]
    \item {\textit{distilbert-base-cased-distilled-squad\footnote{\href{https://huggingface.co/distilbert-base-cased-distilled-squad}{Huggingface: ditsilbert-base-cased-distilled-squad}}}}
    \item \textit{roberta-base-squad2\footnote{\href{https://huggingface.co/deepset/roberta-base-squad2}{Huggingface: roberta-base-squad2}}}
\end{itemize}
\begin{figure}
    \centering
    \includegraphics[width=\columnwidth]{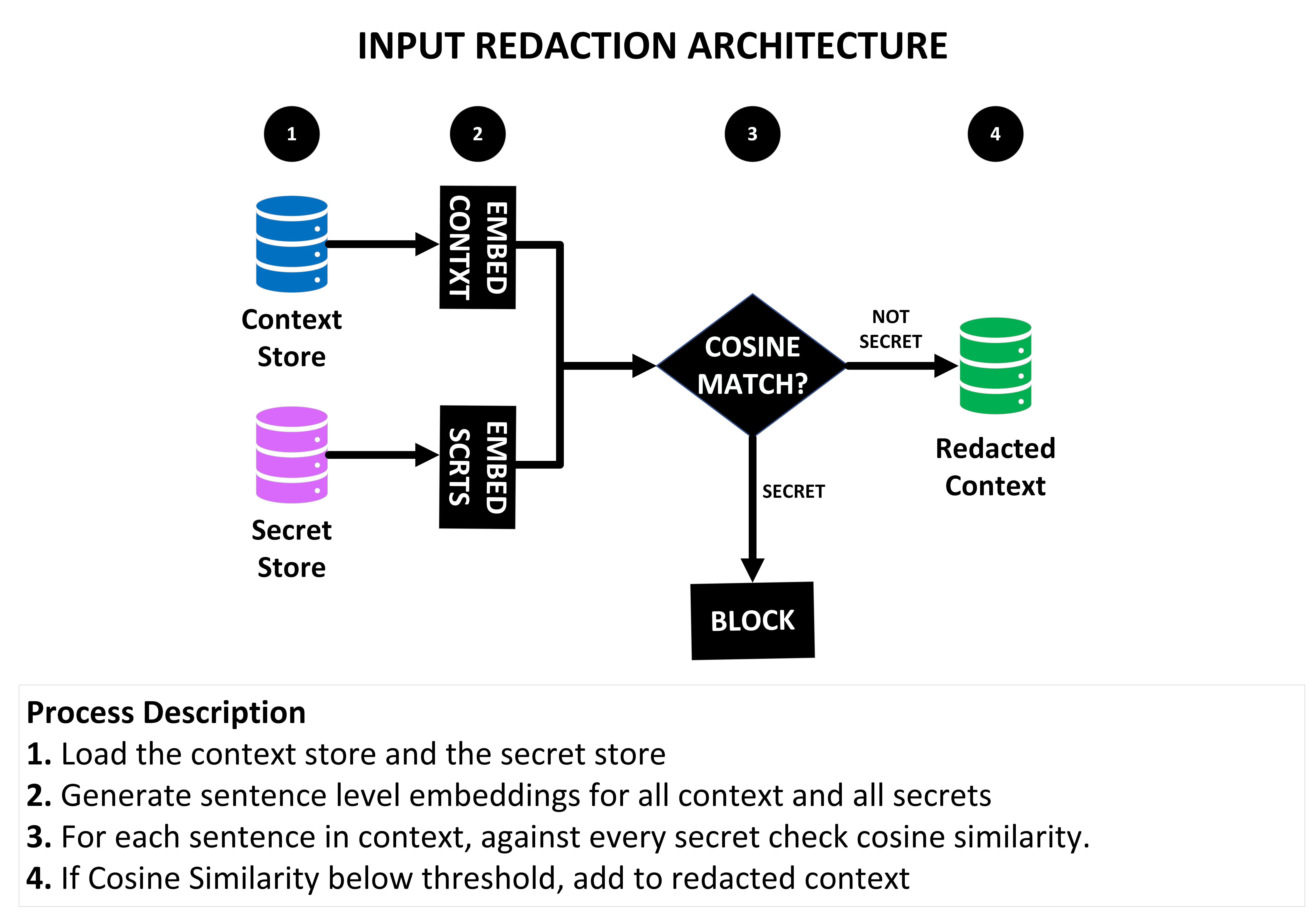}
    \caption{Redaction architecture. Secrets are identified and removed from the data store prior to access by the QA system.  At runtime, the QA system interacts directly with this clean data store.}
    \label{fig:redaction}
\end{figure}

\subsection{Phase 2. Redacted Context Assessment}\label{exp_redacted}
The purpose of our redacted context assessment is to understand how severe the information loss is should destructive redaction to remove secrets occur. The input questions and secrets are unchanged from Phase 1, and we reuse the roberta-base-squad2 model for the QA sytsem. We implement aggressive redaction of secret material from the context in a pre-processing phase. The pre-processing work involves generating sentence-level embeddings for both the secret context and the full SQUAD Dev set context. Unlike the sanitization approach, this redaction must make some assumptions about the size and structure of potential secrets because it has no question and QA model to identify potential question answers. Those embeddings are compared, and only sentences from the context that do not have a corresponding embedding in the secret context are added to the redacted context store. After the pre-processing is complete, the user can ask questions drawing answers from the redacted context. The process is depicted in Figure \ref{fig:redaction}

\subsection{Phase 3. Secret Keeping} \label{exp_sanitize}
\begin{figure}
    \centering
    \includegraphics[width=\columnwidth]{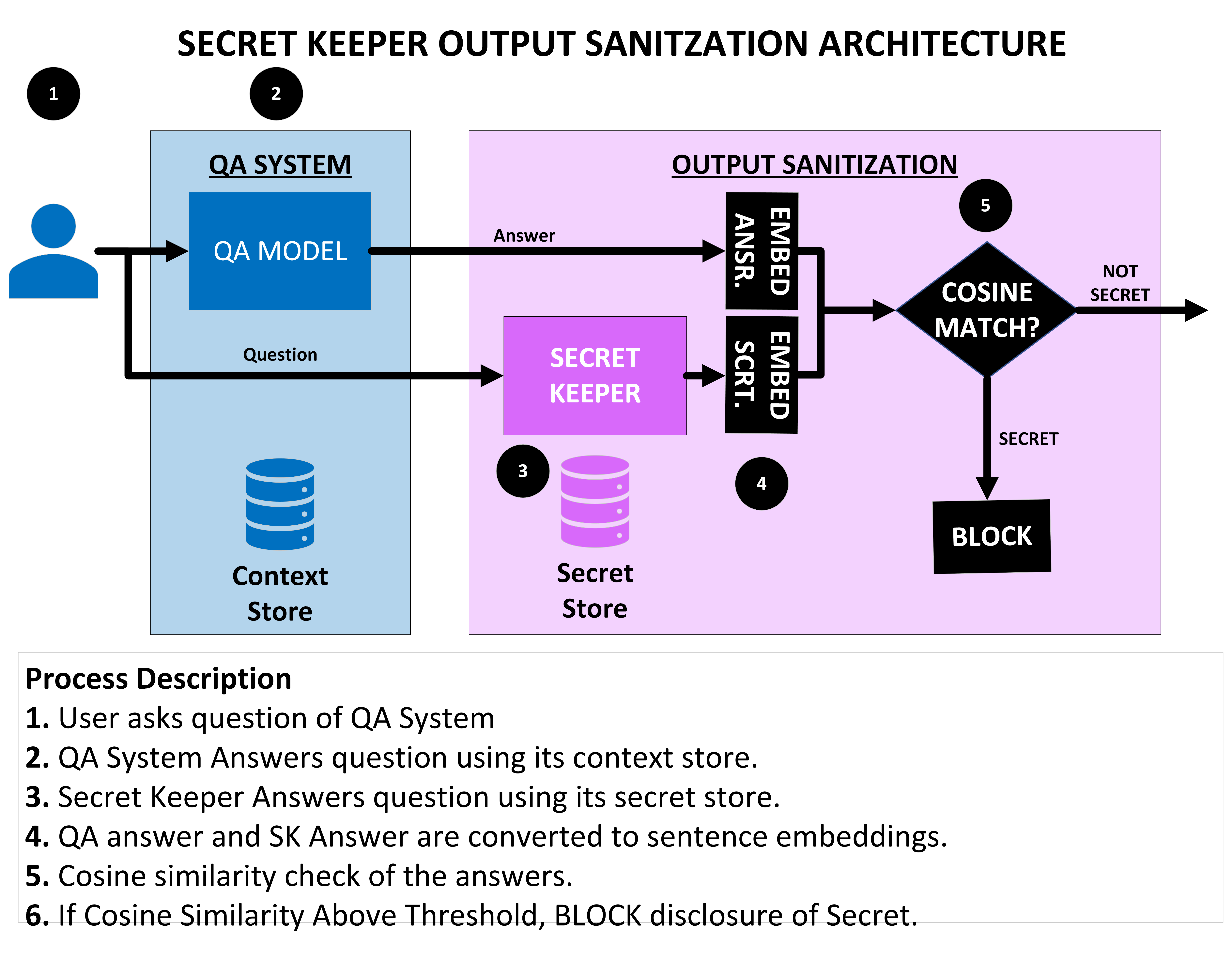}
    \caption{The output-sanitization system architecture.  Both the question-answering and secret-keeping subsystems receive the question and employ a QA model to generate answers.  The secret-keeping system has access to a data store containing only secret information, while the question-answering system has access to the full (secret and non-secret) data.  The results are passed through a sentence encoder, and the cosine similarity between the embeddings is then compared.  If over a threshold determined by the user risk profile, the question-answering subsystem's result is flagged as secret and is not returned to the user.}
    \label{fig:sanitize}
\end{figure}

The purpose of the secret-keeping assessment is to understand the tradeoff between information leakage, paranoia and QA accuracy that our proposed secret-keeping model offers. The input questions, contexts and secrets are replicated from Phase 1. The secret keeping is implemented as an output sanitization function following the QA module of the system. The process is outlined in figure \ref{fig:sanitize}.

To compare the models in the three phases, we conducted our experiment by adjusting three key settings to explore the associated hypotheses:

\begin{table*}[h!]
\resizebox{\textwidth}{!}{
\begin{tabular}{ c|c|c|c|c|c|c } 
\hline
Design & Model & Samples & Accuracy $\uparrow$ & Paranoia $\downarrow$ & Leakage $\downarrow$ \\
\hline
Baseline & RoBERTa & 27,000 & 0.91  & 0.00 & 0.43 \\ 
Baseline & DistilBERT & 27,000 & 0.86  & 0.00 & 0.43 \\ 
Sanitization & RoBERTa & 27,600 & 0.65 & 0.03 & 0.25 \\ 
Sanitization & DistilBERT & 27,000 & 0.67 & 0.03 & 0.26 \\ 
Sanitization & Custom Model & 27,250 & 0.65  & 0.04 & 0.23 \\ 
Secret Remover & RoBERTa & 27,250 & 0.67 & 0.00 & 0.26 \\ 
\end{tabular}}
\caption{Macro Results show \textit{Secret Remover} is the most accurate non-baseline model and minimizes paranoia, but the fine-tuned output-sanitization model minimizes secret leakage. The baselines set the upper bound on accuracy for the sanitization and secret remover systems employing these QA models. Removing secrets necessitates that some correct answers are no longer available and results in lower accuracy.  However, they are unable to identify secrets and therefore provide any secret information requested by the user.}
\label{Table}
\end{table*}

\textbf{H1 - As the number of secrets kept increases, paranoia rates will increase and QA accuracy will decrease.} To test our theory, we varied the number of passages to flag as ’secret’ between 0 and 32. In the Acme scenario, this represents the amount of information they mark as secret, with higher values representing more topics in their system being protected.\\
\textbf{H2 - As the amount of context available to the Secret Keeper decreases, information leakage will increase}. The real world has incomplete and changing contextual knowledge. To simulate these conditions, we varied the percentage of the text from a secret we provided to the secret keeper between 25\% and 100\%, expecting its performance to degrade as the amount of contextual information decreases. While Acme may be able to mark all secret information as such in our previous scenario, this will not always be the case.  Some documents may slip past their initial review, and more documents may be added without updating the secret data store.  The context ratio here represents how much of the information on these secrets have been marked as such, and therefore is accessible to the secret keeper.\\
\textbf{H3 - As the number of questions about secrets increases, information leakage will increase}. To simulate varying intensities of questioning, we change the ratio of questions about secrets in the evaluation set. Our limited proxy implementation of interrogation used randomly selected questions over the secret context, but a more likely real-world scenario would involve targeted questions related to the topic of interest.  However, this approach is intended to provide some insights into how the model behaves as the focus of questions becomes more tightly centred on secrets.  We expect the model's accuracy to approach that of the underlying reader QA system as we decrease the focus on secrets.  Conversely, we expect the simulation of an attacker homing in on the secret by increasing the rate of questions to result in overall accuracy and an increase in leakage.  This approach represents increasingly aggressive questioning by Wile E. Coyote as he tries to get secret information about the failures of Acme products (which they have marked as secret).  \\

Our experiment framework for the number of secrets, context ratio, and question ratio was applied to the three models we implemented into the output-sanitization architecture and the single model implemented in the secret removal architecture. This process yielded a total of 325 discrete experiments, generating over 190,000 question attempts in our evaluation set.

\section{Evaluating the effectiveness of secret-keeping}\label{evaluation}


The evaluation of these experiments consists of a quantitative evaluation of runtimes and overall accuracy, paranoia, and leakage of each model as well as a qualitative investigation of failure cases to gain a deeper understanding of weaknesses in these approaches to the secret-keeping problem.

\begin{figure*}[h!]
\centering

\begin{subfigure}[t]{.49\textwidth}
\includegraphics[width=\textwidth]{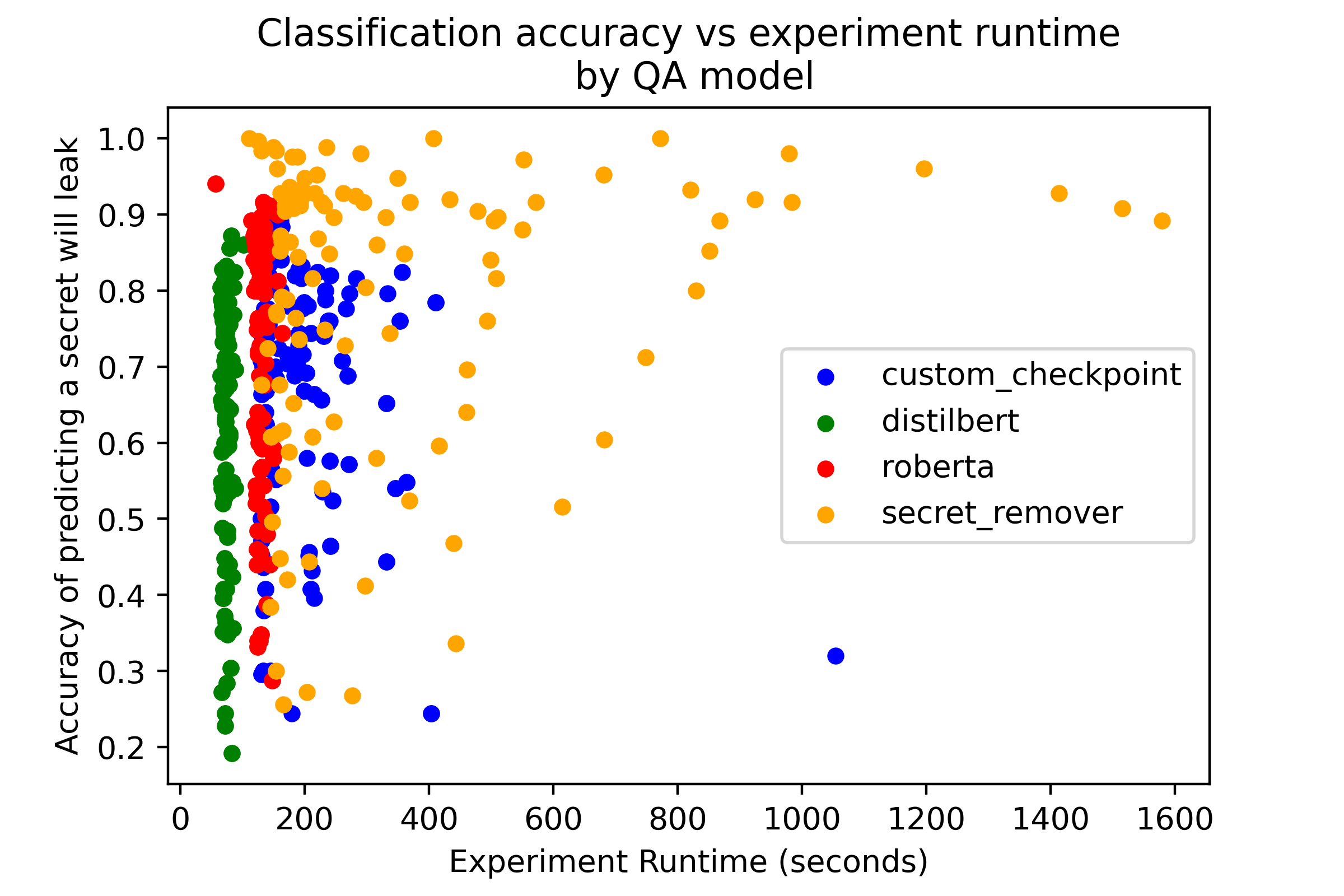}
\caption{Runtime and accuracy by model}
\label{fig:model_comparison}
\end{subfigure}
\hfill
\begin{subfigure}[t]{.49\textwidth}
\includegraphics[width=\textwidth]{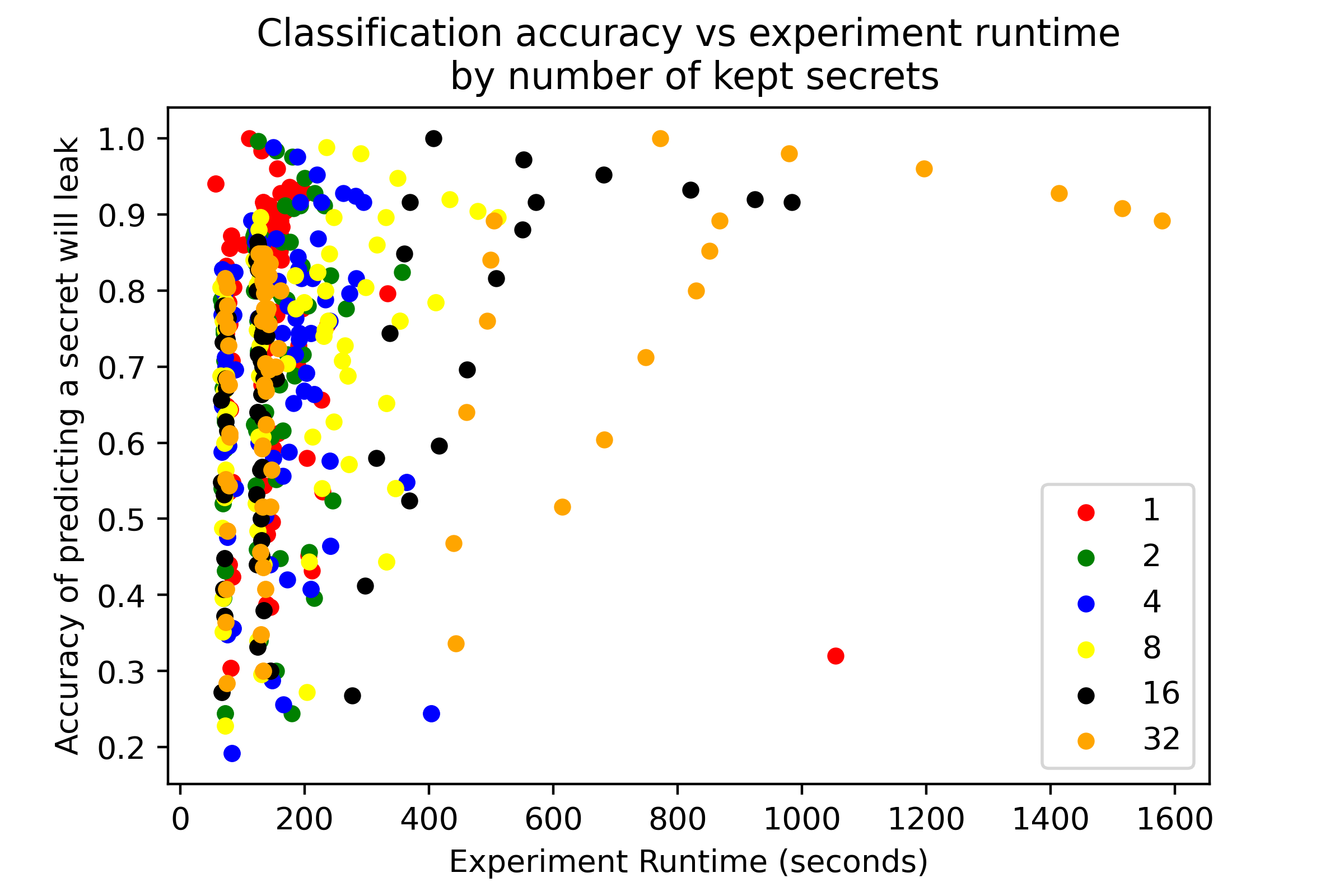}
\caption{Runtime and accuracy by number of secrets}
\label{fig:model_comparison_numSecrets}
\end{subfigure}

\caption{Comparison of model runtime and accuracy.  We see that having more secrets to protect degrades the runtime, but not accuracy, of secretRemover and customCheckpoint, but the other output-sanitization models remain unaffected.}
\label{fig:model_comparison_both}
\end{figure*}

\subsection{Quantitative Results}

Adding a secret keeper to a model along with a secret context will reduce the system's overall question-answering accuracy since the intent is to not return some correct answers as they are secret.  As a result, we see in Table \ref{Table} that the accuracy of each secret-keeping approach is notably lower than that of the base QA system without a secret keeper.  We also observe a substantial decrease in leakage when using all of these models instead of the base system, but the output-sanitization models also introduce paranoia not present in the base models -- they are unable to identify secrets, so they will never mistakenly classify a non-secret as one.\\

Our results supported our hypothesis to varying extents: \\



\begin{figure*}[]
\centering

\begin{subfigure}[t]{.45\textwidth}
\includegraphics[width=\textwidth]{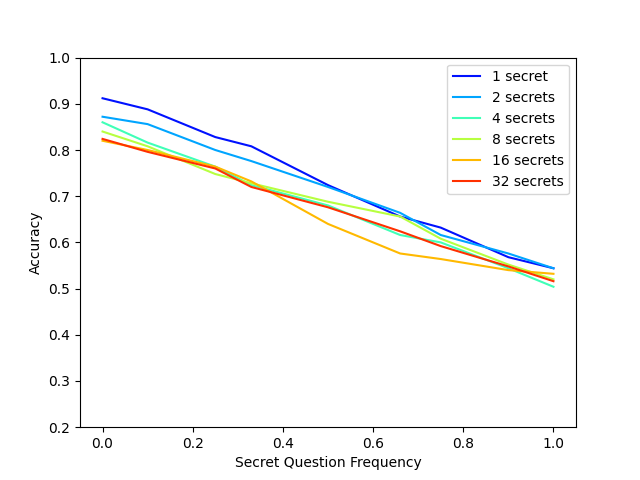}
\caption{Output-Sanitization RoBERTa with 0.5 secret context}
\label{fig:accuracy_vs_secret_ratio}
\end{subfigure}
\hfill
\begin{subfigure}[t]{.45\textwidth}
\includegraphics[width=\textwidth]{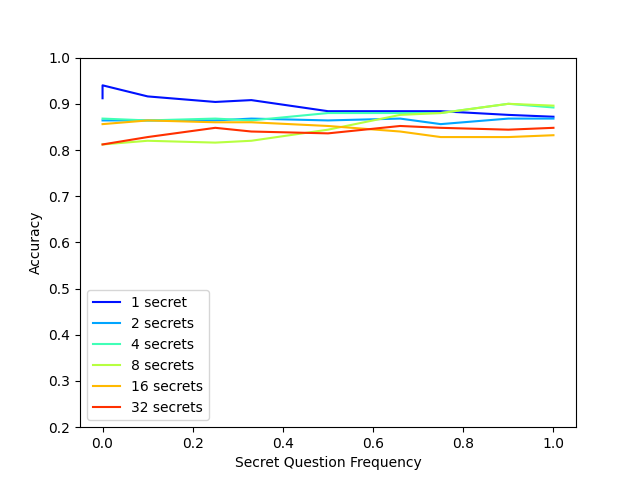}
\caption{Output-Sanitization RoBERTa with 1.0 secret context}
\label{fig:roberta-full-context-f1}
\end{subfigure}

\caption{Results using the RoBERTa Output-Sanitization system with varying secret context coverage. Accuracy in the output-sanitization secret model decreases linearly as the frequency of questions touching on the secret increases with partial context but remains consistent as the secret question frequency increases when provided access to the full secret context.}
\label{fig:roberta_both}
\end{figure*}

\begin{figure*}[h]
\centering

\begin{subfigure}[t]{.3\textwidth}
\includegraphics[width=\textwidth]{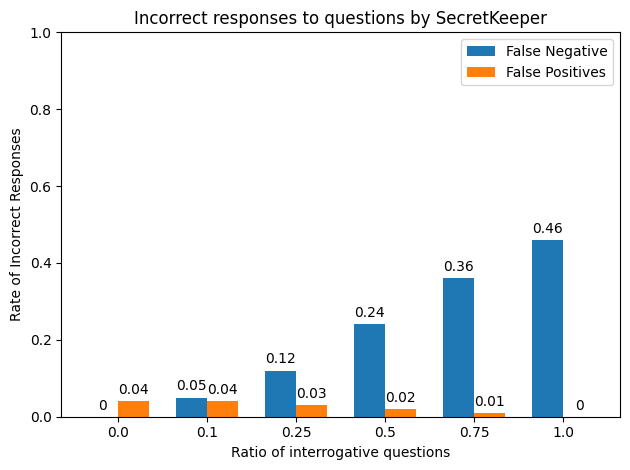}
\caption{Failure cases when varying rate of interrogative questions. The increasing focus on questions targeting secret information drives down false positives (paranoia) across models but results in dramatic increases in false negatives (leakage).}
\label{fig:incorrect_interrogation}
\end{subfigure}
\hfill
\begin{subfigure}[t]{.3\textwidth}
\includegraphics[width=\textwidth]{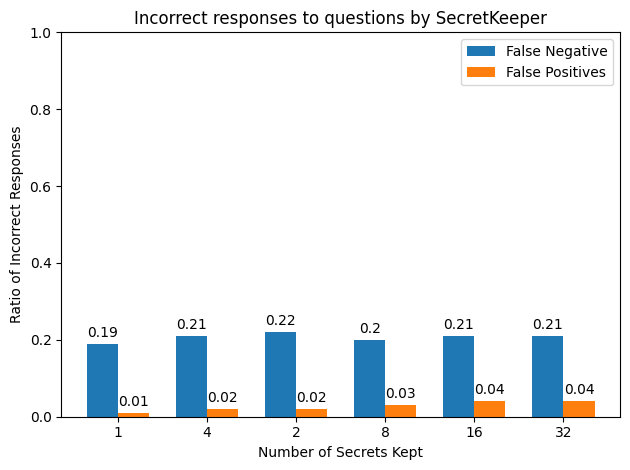}
\caption{Failure cases when varying the number of secrets kept. Increasing the number of secrets kept drives a small but consistent increase in false positives (paranoia).}
\label{fig:incorrect_num_secrets}
\end{subfigure}
\hfill
\begin{subfigure}[t]{.3\textwidth}
\includegraphics[width=\textwidth]{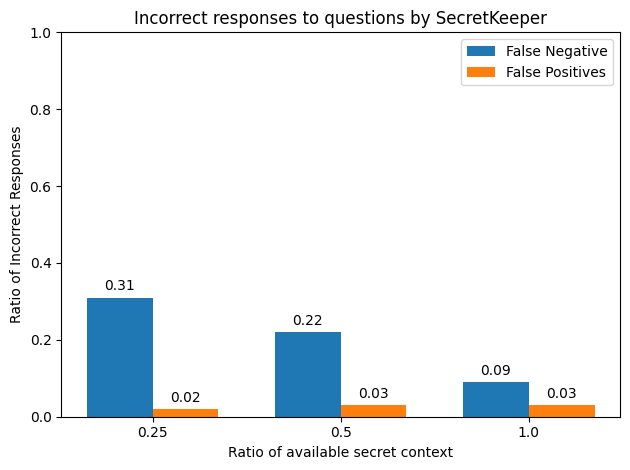}
\caption{Failure cases when varying the amount of available secret context. Increasing the amount of secret context available to models improves performance, but we still see some generalization with only a quarter of the secret context provided to the system.}
\label{fig:incorrect_context}
\hfill
\end{subfigure}

\caption{Failure cases showing the impact of each variable.  All results are averaged across all models for every other combination of variables.}
\label{fig:incorrect_measures_all}
\end{figure*}

 \textbf{H1: Keeping many secrets causes (mild) paranoia.} We varied the number of secrets that the system was required to keep between 1 and 32 to test H1. Our results in Figure \ref{fig:incorrect_num_secrets} show a weak correlation in the number of secrets kept to the occurrence of false positives, steadily increasing from 0.01 to 0.04 as the number of secrets increases.  Figure \ref{fig:roberta-full-context-f1} shows that this effect is most evident when the majority of questions do not target a secret and false positives are more likely. Our qualitative analysis in \ref{qual evaluation} suggests that this phenomenon is a result of our cosine similarity metric relating numerical words, dates, or names with similar patterns together. The exception is the \textit{secret remover} which has a false positive rate of zero.  We attribute the lack of a false positive to its use of full sentences in its embeddings rather than just the phrase returned by a QA model.\\

\textbf{H2: A lack of context causes information leakage}. 
As we varied the amount of secret context available to the secret keeper, simulating keeping secrets on domains with incomplete knowledge. Our results in Figure \ref{fig:incorrect_context} show that when the secret keeper doesn't have access to all the information it needs to explicitly keep secret, its ability to identify secrets across the domain decreases.  However, we do see some generalization as access only to a quarter of the secret context still allows nearly 70\% of secrets to be identified, and full access to the secret context still fails to catch nearly 10\% of the secrets returned by the QA system due to the selection of incorrect secret answers. \\

 \textbf{H3: Interrogation causes information leakage}
 Our experiment varied the ratio of questions asked that were attempting to compromise the secret to simulate what we call the \textit{interrogation mode}. Figures \ref{fig:accuracy_vs_secret_ratio} and \ref{fig:incorrect_interrogation} show when asked more questions pursuing a secret, the probability of leaking the secret increases substantially.  Increasingly aggressive interrogation has a much more pronounced impact than that observed in other experimental changes, as shown in Figures \ref{fig:incorrect_context} and \ref{fig:incorrect_num_secrets}. At the most extreme, nearly half of all secrets were leaked.  This experiment opens the door for developing an adversarial question set to conduct \textit{interrogation} of a secret and provide a basis for testing more advanced systems. \\

We also found some additional valuable results across the models.  While these were not directly related to our original hypotheses, they provide useful insights into comparisons between the different approaches.

\textbf{The secret removal approach scales poorly with more secrets while the output-sanitization models see consistent performance}. Considering Figure \ref{fig:model_comparison_both}, we see this method takes up to 16 times longer than the output-sanitization models as the number of secrets increases. Its performance degrades because the secret removal system must compute embeddings for all sentences in the secret context and all sentences in the QA context and then conduct comparisons on the cosine similarity of all combinations of those embeddings. While it only has to do this once, the time requirement grows significantly as more secrets are added. \\

\textbf{The underlying model drives performance for the output-sanitization designs.} From Figure \ref{fig:model_comparison_both}, we can conclude the runtime performance is heavily dependent on the underlying model for the \textit{output sanitization architecture}. Two of our three models were pre-trained with no modification, and both outperformed the runtime of the fine-tuned model and showed minimal variation as the number of secrets increased.  While fine-tuning remains a viable approach and may provide valuable improvements with additional refinement, our initial experimentation reveals that unmodified pre-trained models perform remarkably well on this task.\\

\textbf{The secret remover is more accurate than the output-sanitization systems.} As shown in Table \ref{Table}, the secret remover has higher accuracy and lower paranoia than any of the output-sanitization systems. However, it struggles with leakage, and its time requirements and more fragile design present additional concerns. In particular, it relies on secrets being sentences, an assumption that does not hold true outside of this experiment and one that is not required by the output-sanitization secret-keeping approach. While it could use other structures instead of sentences, it relies on comparing only these structures, unlike the more flexible QA systems which can select answers of varying lengths and formats based on the question asked and the context supplied.  Additionally, adding new secrets requires further expensive computation and declassification becomes impossible as currently designed because the information no longer exists. Finally, in cases where the original QA context for an answer contained both secrets and non-secrets, this approach may remove critical contextual information for identifying a correct but non-secret answer.\\





\subsection{Qualitative Evaluation} \label{qual evaluation}


Our investigation of the failure cases identified a number of intriguing scenarios that provide some insights into the causes of these failure modes. \\

\example A\Q {What mountain has snow on it all year round? (With secret passage on Kenya)}\QA {Mount Kenya}\SK {Mount Wilson}\R {No Match} B
\textbf{Example 1: Secrets leak when the secret keeper gets the answer wrong:} Occasionally, the secret keeper model will select an incorrect secret answer from the secret context while the QA system finds the correct answer in its context that is also a secret.  As a result, the QA answer, which reveals secret information, slips through undetected. As seen here, this may happen when the QA answer is heavily dependent on the specific context provided.  The secret keeper model is not wrong: Mount Wilson does have snow on it all year, but so does Mount Kenya.  When multiple secret contexts may provide correct answers to a question, this system is likely to make mistakes.  Additionally, it highlights that the underlying model used for the secret QA provides an upper bound on the ability to prevent leakage.  If this model selects incorrect answers, it may not be able to select a valid answer from the secret context in order to compare it against the original QA system's output.\\

\example A\Q {How many weight rooms are in the Malkin Athletic Center (With secret passage on Chemistry)}\QA {three}\SK {two}\R {Match} B
\textbf{Example 2: Numbers and dates cause paranoia (false positives):} Numeric responses always produce high similarity, even if there is little overlap in the original context, causing false positives.  \\ 

\example A\Q {What was the name of the Florida Huguenot colony? (With secret passage on Jacksonville Beaches)}\QA {Fort Caroline}\SK {Fort Caroline}\R {Match} B
\textbf{Example 3: Domain collision can cause paranoia (false positives):} The above is an example of when both models work correctly, but the contexts have unrelated matching information. The secret keeper is drawing from a context about the area around Jacksonville, while the QA model is drawing from a context discussing Huguenots, which have overlapping information since some of these Huguenots settled near Jacksonville. In this case, providing this answer as the name of the colony in the context of a question about the Huguenots rather than about Jacksonville would not have revealed secret information, causing a false positive. \\

\example A\Q {What type of medicine did Otachi focus on? (With secret passage on Pharmacists)}\QA {herbal remedies}\SK {herbal medicine}\R {Match} B
\textbf{Example 4: Differing contexts may determine if the same answer reveals a secret and sometimes leads to paranoia (false positives):} Again, both models work correctly, but the coincidental similarity in the contexts produces a false positive. The QA model was drawing from a context on the Yuan dynasty and a well-known doctor at the time, Otachi. The secret context is on pharmacists. However, one listed specialization of pharmacists is herbal medicine.  Unlike the previous example, both the QA and secret models are referencing the same entity (herbal medicine/remedies), but the difference in source contexts separates the secret related to pharmacists from a historical reference to a doctor in China.  \\ 

This analysis highlights some key weaknesses of this approach, but it also suggests some avenues for improvement.  First, tracking entities within the context around the QA and secret keeper answers may help resolve the issues highlighted in Examples 2 and 3.  If we are able to confirm that the "two" response secret keeper is referring to chemical compounds and the "three" from the QA system refers to the number of rooms in an athletic center, then we may choose to return the QA result despite the high cosine similarity.  However, this approach also introduces some additional concerns as we ultimately care about what the user can learn from the answer.  If a user asked the question "What is the name of a beach in Florida?" with the QA system still returning Fort Caroline (from the colony context) as an answer, the user would still be in possession of a secret despite the accidental domain collision.  In our experiments, the questions were tied to relevant contexts as part of the dataset, but in the wild, that may not always be the case.  Consideration not only of the context of the QA and secret keeper answers but also the question being asked will be critical in further improving these models.

\section{Is it right for machines to keep secrets?} \label{ethics}
Is it ethical for a QA system to withhold information from a user, or lie to them? A strong argument exists that the good stemming from enforcing a requirement for honest and truthful AI far outweighs any potential harms caused by lies or negligent falsehoods \citep{Evans2021}. Reflecting on our motivating example, the question of whether it is more harmful for Acme to release the components and assembly instructions for the \textit{Acme Self-Guided Aerial Bomb} or for a computer to deliberately withhold that information from a user suggests that there are some scenarios where withholding of information may be in the 'right' thing to do. Proceeding under the premise that there exist circumstances where information should be secret, what is the best method?

\textbf{Omission versus Lying} Neither humans nor computers are particularly effective at detecting deception \citep{Peskov2020}, but humans tend to be better at detecting lies that can be verified by checking another source. Obvious omission is suspicious, but covert omission is more effective than the use of outright lies \citep{VanSwol2012}. The effectiveness of omission over lying about the answer is likely linked to the ability to fact-check incorrect statements but the inability to fact-check that which is unknown. From an ethical perspective, the use of omission or glomarisation to conceal information is seen as less problematic than outright lying \citep{Evans2021}. While out of scope for this paper, we identify significant future work in approaches of \textit{satisficing} or \textit{glomarization} to concealing information from a user. Preventing a user from commencing interrogative interactions with the system increases the overall security of information and ameliorates some ethical concerns about active deception.

\textbf{Risks of a Secret File} An obvious risk of our approach to secret-keeping is that the system owner is required to maintain a central repository of all secrets. The repository will be a highly desirable target for attackers. Our current implementation requires the secrets to be in plaintext, but future work should examine how to improve the security of the secret store without unacceptably detracting from the system's performance. 

\section{Related Work in protecting sensitive information} \label{related}
Although the problem of secret keeping in QA systems is new, there are related efforts in QA, agent-based systems, content moderation, LLM Privacy, spoiler detection, censorship, and sanitization that have informed our work. 

\subsection{Question Answering} \label{related_QA}

\textbf{Recent work in extractive question answering focuses on answerability, not protecting secrets.} We examined modern QA datasets including Comqa \citep{Abujabal2018}, HotpotQA \citep{Yang2018}, and Natural Questions \citep{Kwiatkowski2019}. They focus on maximizing the availability of answers rather than preventing answers from being given. Notably, both SQUAD2.0 \citep{Rajpurkar2018} and QuAC \citep{Choi2018} include the notion of \textit{unanswerable questions}. Unfortunately, these are focused on poorly formed questions and do not explore deliberate secret-keeping.

\textbf{The removal of examples from training data reduces the ability of the LLMs to correctly answer questions.} \citet{Kandpal2022} found a correlation between the frequency of examples in training and the likelihood of its generation by the trained model. They also found that the removal of training material significantly degrades the ability of both LLM and retrieval-based QA models to answer factoid questions. 
The negative impact on QA accuracy caused by removing concepts from training data drives our design choice for a system that checks for secret leakage at query-time, rather than during training. 

\subsection{Agent-based systems and reasoning} \label{related_agents}

\textbf{Adversarial inputs could induce the disclosure of strategically sensitive information that undermines an agent's objectives.} A 2022 work by \citeauthor{FAIR2022} develops CICERO, a system that can play the game Diplomacy. Their supplementary material notes that by conditioning on intent they are able to improve the perplexity of generated text \citep{FAIR2022A}. The impact on performance suggests that an understanding of \textit{what} and \textit{why} a piece of information is important and improves the ability to reason about its value. They also found that post-generation filtering of messages sent to other players to be a useful approach in protecting the agent's strategic intentions.  We take a similar approach in checking for secret leaks after the generation of responses.



\subsection{Content Moderation} \label{realted_contentModeration}
\textbf{Traditional content moderation using blacklist keyword matching or 'harmful' or 'not harmful' text classification suffers from domain dependence and lexical fragility.} A 2020 Report from the Transatlantic Working Group, led by \citeauthor{Llanso2020} asserts that the rapid defeat of the Jigsaw Perspective API was due to an inability to handle variations in the lexical inputs. Keyword-based content moderation for the Cicero system had to be supplemented with domain-specific terminology to make it functional \cite{FAIR2022A}. Active learning approaches like that described by \citeauthor{Markov2022} in their 2022 paper showed they are more robust to lexical variances in user behavior attempting to bypass content moderation and warrants further investigation for secret-keeping. 

\subsection{Spoiler Detection} \label{related_spoiler}
\textbf{Spoiler detection is conceptually aligned with secret keeping but does not protect secrets.} Early work in spoiler detection sought to use classification approaches to determine if spans of text are \textit{spoilers} or \textit{not spoilers}. Work by \citeauthor{BoydGraber2013} demonstrated it is possible to detect spoilers through classification. Further work by \citeauthor{Wan2019} in 2019 applies a neural approach and \citeauthor{Chang2021} in 2021 develop a Graph Neural Network that leverages dependency relations to improve their performance. While conceptually similar, spoiler detection is like content moderation in that it tries to detect the language of spoilers as an open-ended problem, rather than focusing on protecting a discrete secret as we propose here. 

\subsection{Memorization and Forgetting} \label{related_memorization}

\textbf{Recent work in memorization by \cite{Jagielski2022} suggests that recently seen training examples are more likely to be memorized, and so fine-tuning an LLM on domain-specific data means it is more likely to leak that domain-specific information}. \textit{Memorization} refers to when a trained neural network reveals the presence of training data. In 2019 \citeauthor{Carlini2019} observed that rare examples can still be memorized and that it is very difficult to detect and prevent unintentional memorization from occurring.  In 2021 \citeauthor{Carlini2021} extended their prior work to show how an attacker can cause the \textit{indiscriminate} extraction of training examples from large language models. That result was further extended by \citeauthor{Tramer2022} in 2022 who showed that by poisoning the training input they are able to increase information leakage using what they term \textit{active inference attacks}. Effective, indiscriminate attacks to leak training examples are very problematic for secret-keeping.  
These approaches focus on 'private' or 'sensitive' categories of information rather than a specific protected secret fact. Additionally, they are effectively side-channel attacks that an interrogator may resort to in an attempt to have a secret disclosed. We focus our secret-keeping system on in-band interactions but note that in a pipeline design like the one we propose, it is possible to detect the likely disclosure of a secret and prevent it, even if the example is memorized by the model.  

\subsection{Censorship, Sanitization and Anonymization}\label{related_censorship}
\textbf{Sanitizing the output of a system is superior to censoring inputs or anonymizing text for preventing information leakage, but the methods available to sanitize outputs are lacking}.  \citeauthor{Mozes2021} assert that current approaches to text anonymization by removing PII from text offer insufficient guarantees of success. Ultimately, their proposed framework to address the guarantee requires human review and as a result, is not scalable enough for general use. \citeauthor{Carlini2019} state that \textit{sanitizing} outputs based on blacklists is unable to handle the many variations of PII, finding that their attempt to implement a sanitizer by training multiple models to sanitize output did not reliably capture all PII. While the sanitization approach may not work for the problem of protecting \textit{any PII}, we believe it is likely to be the most effective approach for protecting \textit{specific secrets} because of the substantially reduced domain variance. We largely derive the idea to use of one QA model to sanitize the output of another from their work, though we implement them in a different manner. \citet{Song2019} suggest \textit{censorship} of inputs as an approach to combat the memorization of training examples in LLMs. They observe that while the memorization of sensitive material decreases, so does the model accuracy. Further, they present a de-censorship attack which is able to reveal some elements of censored input data. Based on their results we contend that allowing the model access to all training material and then effectively sanitizing the output of secret material is a superior approach to censoring input or complete redaction. 

\section{Conclusion and Future Work} \label{conclusion}


We have introduced the task of \textit{secret-keeping} as an important, and under-explored problem in question answering. We identify a lack of suitable secret-keeping metrics and define \textit{secrecy, paranoia} and \textit{information leakage} to address the gap. We design and implement a secret-keeping approach that is model-agnostic, only requiring access to pre-defined secrets, and the output of a QA system to, detect the disclosure of secrets. 

We have identified a rich field for future work in secret-keeping including: 
\begin{itemize}[nosep,labelindent=0pt,itemindent=0pt,leftmargin=*]
    \item \textbf{Reducing Paranoia and Information Leakage.} How do we reduce the rate of paranoia against dates, names and other patterns with high cosine similarity but low semantic similarity?
    \item \textbf{Gold-Standard Dataset}. We need to generate a gold-standard dataset for secret-keeping, or extend current datasets to support secret-keeping to benchmark performance for future efforts in the area. 
    \item \textbf{Other QA Methods.} Our architecture is flexible and so can support any QA answer as input, but formal testing is needed on the ability to handle generative, multi-choice and categorical systems. Categorical in particular will be challenging as yes/no answers will likely require inspection of the question as well.
    \item \textbf{Memorization attacks.} As with Other QA methods, we expect that the results of memorization attacks that disclose secrets can be addressed with secret-keeping, but formal experimentation is required.
    \item \textbf{Resistance to Interrogation} Extending the work by \citeauthor{Markov2022} to apply active learning to reduce the information leakage induced by interrogation.
    \item \textbf{Information Aggregation.} Identifying atomic facts and relationships that are secrets is feasible, but how can a secret-keeper track the \textit{aggregate knowledge} of an interrogator who asks many innocuous questions and combines the answers themselves?
    \item \textbf{Interrogation detection.} How can a QA system detect adversarial patterns of questions that indicate an \textit{interrogation} might be underway? Can the system adjust its strictness of secret-keeping in response to defeat it?
    \item \textbf{Satisficing and Glomarization.} Can the QA system generate answers using techniques like those described by \citeauthor{Evans2021} that will appease them or distract them so that they will not commence an interrogation to help protect information, supporting efforts by \citeauthor{FAIR2022} and \citeauthor{Tabatabaei2023} to protect intent-based systems?
    \item \textbf{Secret Security} How can we protect secrets without storing them in a way that they are vulnerable to direct access by a threat actor? 
\end{itemize}

Our experiments demonstrate that Secret information leaks from unprotected QA systems, and complete redaction of secret material is time-consuming, destructive and still prone to information leakage. Secret-Keeping offers users the ability to trade-off between \textit{paranoia} and \textit{information leakage} without disproportionately sacrificing accuracy in addition to providing extensive flexibility in selecting QA models.

\section{Acknowledgments}
The authors gratefully acknowledge the technical contributions of Ryan Van Voorhis and Ben Moskowitz. 

\bibliography{anthology,SKiQA}
\bibliographystyle{acl_natbib}




\end{document}